\documentclass[a4paper]{article}

\usepackage[english]{babel}
\usepackage[T1]{fontenc}
\usepackage[utf8]{inputenc}
\usepackage{textcomp}
\usepackage[a4paper,top=3cm,bottom=2cm,left=3cm,right=3cm,marginparwidth=1.75cm]{geometry}

\usepackage{amsmath}
\usepackage{graphicx}
\usepackage[colorinlistoftodos]{todonotes}
\usepackage[colorlinks=true, allcolors=blue]{hyperref}
\usepackage{booktabs}
\usepackage{graphicx}
\usepackage{hyperref}
\usepackage{listings}
\usepackage{amsmath}
\usepackage{verbatim}
\usepackage{color}
\usepackage{wrapfig}
\usepackage{lineno}
\usepackage{array}
\usepackage{makecell}
\usepackage{tabularx}
\newcolumntype{Y}{>{\centering\arraybackslash}X}
\usepackage{subcaption}
\usepackage{caption}
\usepackage{listings}
\usepackage{xcolor}
\usepackage{colortbl}
\usepackage{hyperref}
\usepackage{authblk}
\usepackage{longtable}
\usepackage{adjustbox}
\usepackage{multirow}
\usepackage{float}
\usepackage{setspace}
\usepackage{pdfpages}
\usepackage[style=nature,url=true,doi=true,maxbibnames=3, defernumbers=true]{biblatex}
\bibliography{sample.bib}
\usepackage{pdflscape}
\newcolumntype{L}[1]{>{\raggedright\arraybackslash}p{#1}}
\newcolumntype{Y}{>{\raggedright\arraybackslash}X}

\author [1]{Muhammad Rizwan Awan*}
\author[1]{Volker Pickert}
\author[2]{Waqar Muhammad Ashraf}
\author[1]{Saleh Ali}
\author[1]{Farshid Mahmouditabar}
\author[1]{Shafiq Odhano}

\title{\textbf{Towards Agentic Defect Reasoning: A Graph-Assisted Retrieval Framework for Laser Powder Bed Fusion}}

\affil[1]{Department of Electrical and Electronic Engineering, The Newcastle University}
\affil[2]{The Sargent Centre for Process Systems Engineering, Department of Chemical Engineering, University College London, Torrington Place, London, WC1E 7JE, UK}

\affil[*]{corresponding author: muhammad.awan@newcastle.ac.uk}

\begin{document}

\maketitle

\begin{abstract}
Laser Powder Bed Fusion (LPBF) is highly sensitive to process parameters, which influence defect formation through complex thermal and fluid mechanisms. However, defect-related knowledge is dispersed across the literature, limiting systematic understanding. This study presents a graph-assisted retrieval framework for defect reasoning in LPBF, using Ti–6Al–4V as a case study. Scientific publications are transformed into a structured representation, and relationships between parameters, mechanisms, and defects are encoded into an evidence-linked knowledge graph. The framework integrates semantic and graph-based retrieval, supported by a lightweight agent-based reasoning layer to construct interpretable defect pathways. Evaluation shows high retrieval accuracy (0.9667) and recall (0.9667), demonstrating effective identification of relevant defect-related evidence. The framework enables transparent reasoning chains linking process parameters to defects. This work provides a scalable approach for converting unstructured literature into a query able and interpretable knowledge resource for additive manufacturing.

\end{abstract}

\vspace{1em} 

\section{Introduction}
Laser Powder Bed Fusion (LPBF) is widely used to manufacture complex metallic parts with high geometric freedom and good dimensional control. At the same time, the process is very sensitive to changes in processing conditions. Parameters such as laser power, scan speed, hatch spacing, and layer thickness influence melt-pool behavior, heat flow, and solidification, which then affect the formation of porosity, lack-of-fusion defects, keyhole defects, and cracking \cite{snow2020invited, king2014observation}. Experimental and modeling studies have shown that keyhole formation, vapor depression, recoil pressure, and complex melt flow can directly contribute to pore formation and other defect features in LPBF builds \cite{khairallah2016laser, bayat2019keyhole}. Broader reviews of metal additive manufacturing have also shown that process, structure, and properties are strongly coupled, which makes defect control a central challenge in reliable part production \cite{debroy2018additive, chowdhury2022laser}.

This challenge is not only physical but also informational. Knowledge about LPBF defects is spread across a large and growing literature. Some studies focus on porosity maps or process windows \cite{gordon2020defect, du2019effects}, while others examine keyhole porosity, lack-of-fusion pores, or residual-stress-related damage in more specific settings \cite{yang2023formation, shrestha2022formation, bastola2023review}. Review papers have helped summarize this knowledge, but the field still lacks a structured way to connect process parameters, defect types, and underlying mechanisms across many publications \cite{bera2024review}. As a result, researchers often have to manually search, compare, and interpret information that is reported with different terminology and different levels of detail.

Materials informatics and natural language processing offer a possible way forward. Prior work has shown that scientific literature can be mined to extract structured knowledge from large document collections \cite{kononova2019text, venugopal2021looking, shetty2021automated}. More recent studies have shown that large language models can support structured information extraction from scientific text and can recover materials data from research papers with high accuracy when guided carefully \cite{shetty2021automated, zaki2022extracting}. These studies are encouraging, but they are not focused on LPBF defect reasoning, and they do not by themselves provide a framework for tracing how parameters, defects, and mechanisms are linked in additive manufacturing literature.

In parallel, retrieval-augmented generation has become an important way to ground language-model responses in retrieved evidence rather than relying only on model memory \cite{lewis2020retrieval, polak2024extracting, dagdelen2024structured}. This is relevant in technical domains because unsupported answers and hallucinated reasoning remain a known concern in large language models \cite{procko2024graph}. However, standard retrieval methods are usually based on semantic similarity alone. That is useful for finding relevant passages, but it is less effective when the task requires multi-step reasoning over linked concepts such as a process parameter, a defect type, and a mechanism. Graph-based retrieval has therefore attracted attention as a way to preserve and use relationships between entities during retrieval and answer generation \cite{edge2024local, peng2025graph}.

Despite these developments, there is still a clear gap in LPBF defect analysis. Existing LPBF studies provide rich defect knowledge \cite{snow2020invited,gordon2020defect, yang2023formation}, materials-text-mining studies provide methods for extracting structured information from literature \cite{kononova2019text, zaki2022extracting}, and RAG studies provide methods for evidence-grounded generation \cite{polak2024extracting, edge2024local}. What is still missing is a focused framework that brings these strands together for defect reasoning in LPBF . In particular, there is a need for an approach that can turn a corpus of LPBF publications into a structured knowledge resource and then use that resource to support defect-focused question answering and interpretable reasoning.

This work addresses that gap by developing a literature-grounded framework for LPBF defect analysis. The study focuses on a corpus of open-access publications related to a single material system, namely the titanium alloy Ti–6Al–4V, which is used as a representative case study in this research. The framework combines structured extraction from cleaned literature, graph-based representation of extracted relationships, and retrieval-augmented answering, supported by a lightweight agentic reasoning layer. The objective of this study is to demonstrate how open scientific literature can be systematically transformed into a structured and queryable knowledge resource for defect-focused reasoning in additive manufacturing, using a controlled material-specific case study.

\section{Methodology}
\subsection{Data collection and preprocessing}

\begin{figure}[htp]
    \centering
    \includegraphics[width=1.0\linewidth]{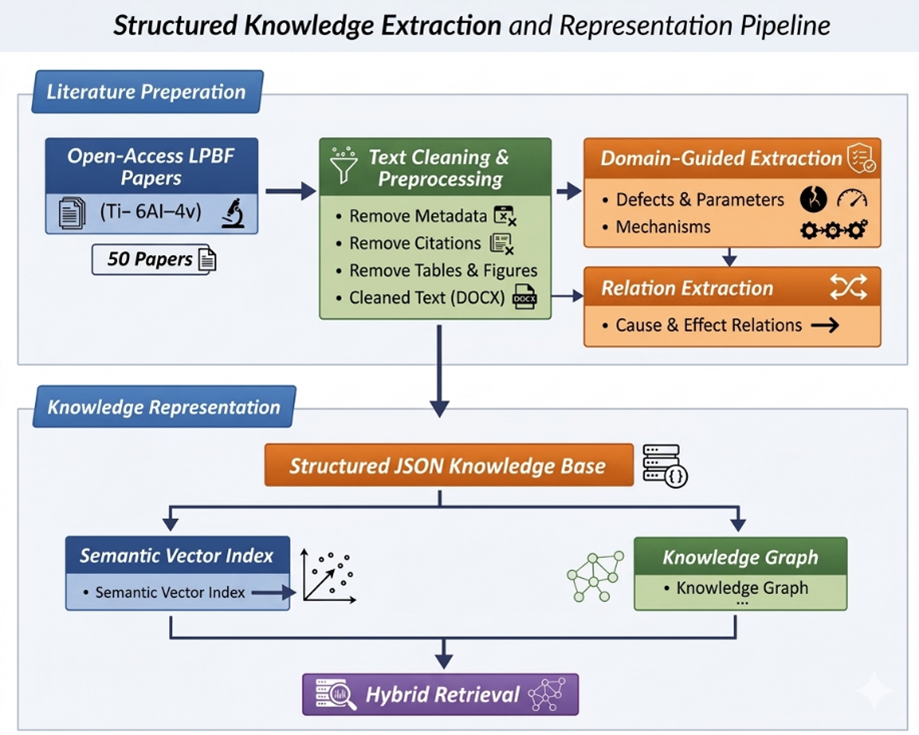}
    \caption{Data pre-processing Pipeline.}
    \label{Fig:fig_1}
\end{figure}

The data collection pre-processing pipeline has been shown in Figure \ref{Fig:fig_1}. A focused literature corpus was constructed consisting of 50 open-access research publications related to Laser Powder Bed Fusion (LPBF) of Ti–6Al–4V. The selection of a single alloy system was intentional, allowing the study to isolate defect–process relationships within a controlled material context and avoid variability introduced by different material systems.

The collected documents were pre-processed to remove non-informative elements including metadata, references, citations, tables, and figure descriptions. This step ensured that only narrative technical content was retained for downstream analysis. The preprocessing pipeline operated on cleaned DOCX files, where textual content was extracted using python-docx. Text normalization was performed using standard Python utilities, including regular expression-based cleaning (re) to remove formatting artifacts and normalize whitespace. Sentence segmentation was carried out using the NLTK tokenizer to preserve semantic boundaries during subsequent chunking.

This preprocessing stage resulted in a consistent and noise-reduced textual corpus suitable for structured extraction and retrieval.

\subsection{Text segmentation and structured}

To enable efficient retrieval and localized reasoning, each document was segmented into overlapping textual units. A sentence-aware chunking strategy was adopted, where text was divided into segments of approximately 220 words, with an overlap of 40 words between consecutive chunks. This approach preserved contextual continuity across chunk boundaries while maintaining manageable input sizes for embedding and language model processing.

Each chunk was assigned a unique identifier and associated with metadata including document source, position index, and word count. The segmented corpus was then transformed into a structured JSON format, where each chunk served as the fundamental unit of representation.
For each chunk, domain-relevant information was extracted and stored under predefined fields, including:
\begin{itemize}
    \item defect-related terms 
	\item process parameters 
	\item mechanisms 
	\item consequences 
	\item entities and relations 
\end{itemize}

The resulting JSON structure provided a unified representation that supported both semantic retrieval and graph construction. This representation ensured that all extracted knowledge remained traceable to its originating text segment.

\subsection{Domain-guided knowledge extraction}

Knowledge extraction was performed using a domain-guided, rule-based approach. Controlled vocabularies were defined for key LPBF concepts, including defects, process parameters, physical mechanisms, and process descriptors. These vocabularies were used to identify relevant terms within each chunk.

Relation extraction was conducted at the sentence level to capture localized causal and associative statements. Linguistic patterns were used to identify relationships such as:

\begin{itemize}
    \item causal links (e.g., “leads to”, “results in”) 
	\item increasing or decreasing effects 
	\item general influence relationships 
\end{itemize}

Based on detected patterns, relationships were constructed between identified entities, including:
\begin{itemize}
    \item parameter–defect relationships 
	\item parameter–mechanism relationships 
	\item mechanism–defect relationships 
	\item defect–consequence relationships 
\end{itemize}

These relations were encoded as structured triples and stored alongside each chunk. The use of rule-based extraction ensured interpretability and alignment with domain knowledge, while maintaining consistency across the corpus.

\subsection{Knowledge graph construction}

\begin{figure}[htp]
    \centering
    \includegraphics[width=1.0\linewidth]{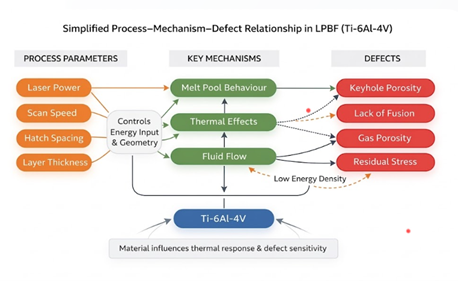}
    \caption{Knowledge graph construction.}
    \label{Fig:fig_2}
\end{figure}

A directed knowledge graph was constructed from the extracted relational data, which links material, defects, parameters and associated mechanisms as shown in Figure \ref{Fig:fig_2}. In this representation, nodes correspond to normalized domain entities such as process parameters, defects, and mechanisms, while edges represent directional relationships derived from the literature.

When multiple pieces of evidence supported the same relationship, edge weights were incremented and supporting chunk identifiers were aggregated. This allowed each edge to retain a link to its originating textual evidence, enabling traceability between the graph structure and the literature corpus.

The knowledge graph therefore served a dual purpose:
\begin{itemize}
    \item representing structured relationships between entities 
	\item acting as an index linking concepts to supporting evidence 
\end{itemize}

This evidence-linked graph structure enabled graph-guided retrieval while preserving the original textual grounding of extracted knowledge.

\subsection{Semantic indexing and text-based retrieval}
In parallel with graph construction, a semantic retrieval system was developed using sentence embeddings. Chunk-level embeddings were generated using the Sentence Transformers model all-MiniLM-L6-v2. These embeddings were indexed using FAISS, enabling efficient similarity-based search over the corpus.

Given a user query, its embedding was computed and compared against the indexed chunk embeddings. The top-k most similar chunks were retrieved based on distance in the embedding space. This approach formed the baseline text-based retrieval component of the system.
The retrieved chunks provided candidate evidence for downstream answer generation.

\subsection{Graph-assisted retrieval}
\begin{figure}[htp]
    \centering
    \includegraphics[width=1.0\linewidth]{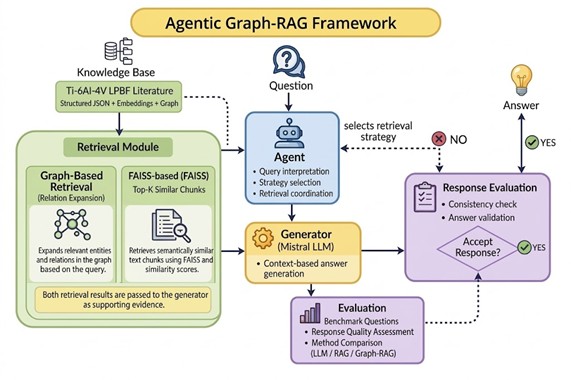}
    \caption{Agent based graph assisted retrieval framework}
    \label{Fig:fig_3}
\end{figure}

To incorporate structured knowledge into retrieval, a graph-assisted retrieval mechanism was introduced as shown in Figure \ref{Fig:fig_3}. User queries were first normalized to align with the graph vocabulary, ensuring consistent matching between query terms and graph nodes.

Relevant nodes were identified from the query, and neighboring nodes were explored through graph traversal. Supporting chunk identifiers associated with these graph connections were then collected to form a candidate evidence set.

If graph-based candidates were available, semantic similarity was computed within this constrained set to identify the most relevant evidence. If no graph matches were found, the system reverted to standard text-based retrieval.

This approach enabled the integration of semantic similarity with literature-derived relational structure, allowing retrieval to be guided by both textual relevance and domain knowledge.

\subsection{Hybrid retrieval and agentic reasoning}
To improve robustness, a hybrid retrieval strategy was implemented, combining graph-based and text-based retrieval outputs as shown in Figure \ref{Fig:fig_3}. Evidence from both sources was merged using a weighted scoring approach, where graph-derived evidence was prioritized over purely semantic matches.

An agentic reasoning layer was introduced to coordinate retrieval and reasoning processes. This layer performed:
\begin{itemize}
    \item query interpretation 
	\item retrieval strategy selection 
	\item evidence aggregation 
	\item reasoning-chain construction 
	\item confidence estimation 
\end{itemize}

Queries were classified into categories such as lookup, comparison, and explanation, and retrieval strategies were selected accordingly. Retrieved evidence was aggregated and analyzed to identify dominant parameters, mechanisms, and defects. A reasoning chain was constructed either from frequent evidence patterns or from graph-based paths connecting relevant entities.

The agentic layer functioned as a coordination mechanism rather than a fully autonomous system, providing structured reasoning outputs grounded in retrieved evidence.

\section{Results and Discussion}
\subsection{Answer generation}
Answer generation was performed using the instruction-tuned language model Mistral-7B-Instruct-v0.2. Retrieved evidence chunks were concatenated into a structured context and provided to the model through a constrained prompt.

The model was instructed to generate answers based solely on the provided evidence. Generation parameters were configured to favor deterministic and concise outputs, with low-temperature sampling.

This ensured that responses remained grounded in retrieved literature rather than relying on the model’s internal knowledge.

\subsection{Evaluation protocol}
The framework was evaluated using a set of 10 benchmark questions designed to reflect common defect-related queries in LPBF. These questions focused on relationships between process parameters, defects, and mechanisms.

Evaluation was primarily conducted at the retrieval level. Retrieved evidence was analyzed to determine whether expected defect-related information was successfully captured. Metrics included:
\begin{itemize}
    \item retrieval accuracy 
	\item precision and recall of defect labels 
	\item parameter identification accuracy 
	\item response latency 
\end{itemize}

This evaluation approach reflects the core objective of the framework, which is to support evidence-grounded reasoning rather than standalone prediction.

\subsection{Knowledge representation and retrieval capability}
The proposed framework successfully transformed unstructured LPBF literature into a structured, query able knowledge resource. The combined JSON and graph representations enabled both semantic and relational access to defect-related information.

The knowledge graph captured meaningful relationships between process parameters, mechanisms, and defects, while maintaining direct links to supporting textual evidence. This allowed the system to operate as an evidence-grounded reasoning framework rather than a purely symbolic model.

\subsection{Benchmark evaluation} \label{examp_1}

To systematically evaluate the performance of the proposed framework, a set of 10 benchmark questions was designed, focusing on defect-related reasoning in LPBF. These questions targeted relationships between process parameters, defect formation, and underlying mechanisms, including porosity, lack of fusion, keyhole behavior, and residual stress.

For each query, the system retrieved a set of top-k evidence chunks. The retrieved evidence was then analyzed to extract predicted defect labels and process parameters. To reduce noise, only labels appearing in at least two retrieved chunks were retained, ensuring that predictions were supported by multiple pieces of evidence.

The evaluation was conducted using standard information retrieval metrics, including:
\begin{itemize}
    \item \textbf{Precision}, defined as the proportion of retrieved labels that are relevant 
    \item \textbf{Recall}, defined as the proportion of relevant labels that are successfully retrieved 
	\item \textbf{F1-score}, representing the harmonic mean of precision and recall 
	\item \textbf{Accuracy}, defined as the proportion of correctly identified defect labels per query
\end{itemize}

\begin{table}[ht]
    \centering
    \caption{Performance Metrics by Question ID}
    \label{tab:metrics}
    \begin{tabular}{llcccc}
        \hline
        \textbf{Question ID} & \textbf{Query Type} & \textbf{Precision} & \textbf{Recall} & \textbf{F1-score} & \textbf{Accuracy} \\ \hline
        Q1 & Explanation & 0.60 & 1.00 & 0.75 & 1.00 \\
        Q2 & Explanation & 0.50 & 1.00 & 0.67 & 1.00 \\
        Q3 & Lookup & 0.67 & 1.00 & 0.80 & 1.00 \\
        Q4 & Explanation & 0.50 & 1.00 & 0.67 & 1.00 \\
        Q5 & Explanation & 0.50 & 1.00 & 0.67 & 1.00 \\
        Q6 & General & 0.60 & 1.00 & 0.75 & 1.00 \\
        Q7 & Explanation & 0.50 & 1.00 & 0.67 & 1.00 \\
        Q8 & Explanation & 0.33 & 1.00 & 0.50 & 1.00 \\
        Q9 & Explanation & 0.60 & 1.00 & 0.75 & 1.00 \\
        Q10 & Explanation & 0.67 & 1.00 & 0.80 & 1.00 \\ \hline
    \end{tabular}
\end{table}

\subsection{Retrieval performance}

The quantitative evaluation implemented in the notebook showed that the Graph-RAG pipeline achieved strong retrieval of defect-related evidence across the benchmark questions. Averaged over the 10-question benchmark, the reported results were:
\begin{itemize}
    \item \textbf{retrieval accuracy:} 0.9667 
	\item \textbf{defect label precision:} 0.6167 
	\item \textbf{defect label recall:} 0.9667 
	\item \textbf{mean latency:} 6.4065 s 
	\item \textbf{mean parameter accuracy:} 0.7222
\end{itemize}
Evaluation on the benchmark questions demonstrated strong retrieval performance. The system consistently identified relevant defect-related evidence, with high recall across queries.

The high recall indicates that relevant defect information was effectively captured within the retrieved evidence. Precision was comparatively lower, reflecting the presence of additional related defect information within retrieved chunks. This behavior is consistent with literature-based retrieval, where individual text segments often contain multiple related concepts.

Parameter identification performance was moderate, suggesting that process parameters were not always explicitly localized within individual chunks. This highlights the distributed nature of parameter information in scientific literature.

\subsection{Agentic reasoning behaviour} \label{exam_3}

The agentic extension demonstrated the ability to coordinate retrieval and reasoning processes. By selecting retrieval strategies based on query type and organizing evidence into structured reasoning chains, the system produced more interpretable outputs.

Confidence estimation provided an additional layer of transparency, allowing responses to be categorized based on the strength of supporting evidence. While this mechanism remains heuristic, it contributes to making the reasoning process more explicit.

\subsection{Qualitative example of defect reasoning} \label{res_exam}

To illustrate the behaviour of the proposed framework, a representative example is presented for the following query:
\textbf{Question:}
\textit{Why does high laser power lead to keyhole porosity in LPBF?}

The aggregated evidence indicated that high laser power increases energy input, resulting in the formation of a deep and unstable keyhole. Collapse of this keyhole can trap vapor within the melt pool, leading to pore formation.

Based on the retrieved evidence, the system constructed the following reasoning chain:

\textbf{laser power} → \textbf{keyhole instability} → \textbf{porosity}

The generated response was consistent with known LPBF defect mechanisms and aligned with the supporting literature. The example demonstrates how the integration of graph-based retrieval and semantic search enables the system to produce interpretable, evidence-grounded explanations.

\section{Conclusion}

This study developed a literature-grounded, graph-assisted retrieval framework for defect-focused reasoning in Laser Powder Bed Fusion. By transforming a corpus of LPBF publications into a structured representation and constructing an evidence-linked knowledge graph, the framework enables systematic exploration of relationships between process parameters, physical mechanisms, and defect formation.

The integration of semantic retrieval and graph-assisted retrieval proved effective in capturing relevant defect-related knowledge, with high recall and strong retrieval accuracy observed across benchmark queries. The framework supports interpretable reasoning chains, allowing defect formation pathways, such as parameter → mechanism → defect—to be explicitly traced back to supporting literature. This addresses a key limitation in existing approaches, where knowledge remains dispersed and difficult to connect.

The simplified knowledge representation introduced in this work further enhances clarity by consolidating complex physical interactions into core mechanism groups, making the framework both interpretable and scalable. The inclusion of a lightweight agentic reasoning layer improves coordination between retrieval and reasoning processes, contributing to more structured and transparent outputs.

Despite these contributions, several limitations remain. The rule-based extraction approach may not capture all nuanced relationships present in the literature, and the evaluation is limited to a relatively small benchmark set and a single material system (Ti–6Al–4V). Additionally, the current reasoning layer does not fully support iterative or multi-hop reasoning beyond predefined patterns.

\section{Limitations}
The current framework has several limitations. The relation extraction process is rule-based and may not capture all relevant relationships present in the literature. The benchmark set is limited in size and scope, and the evaluation focuses primarily on retrieval performance rather than full reasoning accuracy.

The agentic layer remains lightweight and does not incorporate iterative planning or multi-step verification beyond simple feedback loops. Additionally, the framework is evaluated within a single material system, which limits generalizability.

Despite these limitations, the study demonstrates that literature-derived knowledge can be structured and used to support defect-focused reasoning in LPBF.

\section*{Data availability}
The data would be provided upon request.

\section*{Competing interests}
The authors declare no competing interests.

\section*{Research funding}
This research did not receive funding.

\printbibliography

\end{document}